\documentclass[preprint,review,numbers,sort&compress]{elsarticle}
\usepackage[inner=1.0in, outer=1.0in, top=1.0in, bottom=1in, paperheight=11in, paperwidth=8.5in]{geometry}

\usepackage{tikz}
\usepackage[utf8]{inputenc}
\usepackage{amsfonts}
\usepackage[utf8]{inputenc}
\usepackage{dirtytalk}
\usepackage{graphicx}
\usepackage{float}
\usepackage{subcaption}
\graphicspath{{}}
\usepackage{caption}
\usepackage{xcolor}

\usepackage{epsfig} % for postscript graphics files
\usepackage{amsmath} % assumes amsmath package installed
\DeclareMathOperator*{\argmin}{arg\,min}
\usepackage{algorithm}
\usepackage{algpseudocode}

\title{A Bayesian Approach for Shaft Centre Localisation in Journal Bearings}
\author{Christopher A. Lindley, Scott Beamish, Rob Dwyer-Joyce, Nikolaos Dervilis and Keith Worden}
\date{March 2021}

\begin{document}

\begin{abstract}
    It has been shown that ultrasonic techniques work well for online measuring of circumferential oil film thickness profile in journal bearings; unfortunately, they can be limited by their measuring range and unable to capture details of the film all around the bearing circumference. Attempts to model the film thickness over the full range of the bearing rely on deterministic approaches, which assume the observations to be true with absolute certainty. Unaccounted uncertainties of the film thickness may lead to a cascade of inaccurate predictions for subsequent calculations of hydrodynamic parameters. In the present work, a probabilistic framework is proposed to model the film thickness with Gaussian Processes. The results are then used to estimate the location of the bearing shaft under various operational conditions. A further step in the process involves using the newly-constructed dataset to generate likelihood maps displaying the probable location of the shaft centre, given the bearing rotational speed and applied static load. The results offer the possibility to visualise the confidence of the predictions and allow the true location to be found within an area of high probability within the bearing's bore.
\end{abstract}

\maketitle

\section{Introduction}

Traditionally, journal bearing dynamics are modelled theoretically, either by using simplified versions of the Reynolds' equation in one dimension, or with numerical methods to solve for two dimensional cases of both the Reynolds' and energy conservation equations \cite{hamrock1994}. The physical effects typically include the film pressure, film temperatures, and mechanical deformations. It is of importance to monitor these effects to ensure a reliable and safe operation of the bearing. 

Analytically, just knowing the main dimensions and operating parameters of the bearing (e.g. shaft and bore radius, oil temperature and viscosity, loads and rotational speed) is enough to produce deterministic models capable of predicting physical effects under various conditions. In practice, these analytical methods must be accompanied by experimental work to account for complex non-linearities, measurement errors and sources of noise. A common starting point to build models based on experimental observations is to measure the fluid-film thickness of the bearing during its operation. This parameter is usually chosen because of its high dependency on the operating parameters of the bearing. Changing them in any way will adjust the bearing shaft to a new position so that the hydrodynamic forces balance to the applied loads \cite{He2016}. Hence, knowing the fluid-film thickness for a given operating condition allows one to determine some of the hydrodynamic properties of the bearing. A variety of methods have been proposed to measure the oil film thickness of a journal bearing in operation. These have proven to be reliable when based on electrical methods \cite{Cui2014,Glavastskih2001} or optical methods \cite{Fang2017,Marx2016}. Although the predictions of these methods tend to be in good accordance with simulations and numerical solutions, they can be highly complex or rely on impractical requirements (e.g.\ invasive sensor placement or transparent surfaces for optical-based measurements). An alternative implementation based on ultrasonic techniques has the potential to provide robust film thickness measurements without the aforementioned disadvantages. Ultrasonic measurement methods have successfully been demonstrated by extracting features from reflected pulses travelling through a liquid medium between solid surfaces \cite{Zhang2019,DJ2003}. These features are given by the changes in amplitude and phase of the reflected pulses, which encompass the \textit{spring model} for film measurements since these are governed by the stiffness of the medium. The spring model is limited by the range of thicknesses it can measure and an additional model, the \textit{resonant dip technique}, is introduced to account for thicker layers \cite{Hunter2012}. The application of ultrasonic methods on journal bearings has also been studied extensively for online oil-film thickness measurements \cite{DJ2003,Kasonlang2013,Reddyhoff2005,Beamish2021}.

Although the results from these studies were promising, limitations in their methods are evident because the restrictions imposed by the amplitude change method, phase change method and resonant dip technique. This work attempts to overcome some of these limitations by constructing models in a probabilistic framework. Additionally, these models will be used to determine likelihood maps of the shaft-centre position. Before continuing with the modelling process, a brief theoretical background in ultrasonic methods and their limitations is covered.

\section{Background}

\subsection{Ultrasonic technique for oil film thickness}
\label{sec:ultrasound_techniques}

This work refers to the experimental setup used in \cite{Beamish2021}, and the reader is encouraged to review their work for a complete description. Nevertheless, in order to contextualise over the following sections, some key points and main results will be reiterated here.

The rig used comprised of a bespoke journal bearing designed to have a set of ultrasonic emitters and receivers embedded in the shaft. Pulses of ultrasonic waves travel through the oil film as the shaft spins, with some of their energy bouncing off the film and adjacent surfaces. The reflected waves are then picked up by the same emitting sensors and the signals processed to identify possible changes in their features. 

One of these features is the amplitude of the ultrasonic wave, which is expected to experience a reduction given that some of its energy will be absorbed by the fluid-film. The idea is to somehow relate this measured change in amplitude to the actual thickness of the film. This relation is achieved by defining the reflection coefficient $R$, which is given by the ratio of the reflected signal amplitude to the incident signal amplitude, together with acoustic properties of the bearing and lubricant film. The result is an expression,
\begin{equation}
    h = \frac{\rho c^2}{\Omega z_{1} z_{2}}\sqrt{\frac{R^2(z_{2} + z_{1})^2 - (z_{2} - z_{1})^2}{1 - R^2}}
\label{amplitude_change_eq}
\end{equation}
where $h$ is the fluid-film thickness, $\rho$ is the lubricant density, $c$ the velocity of sound in the lubricant, $\Omega$ the angular frequency of the ultrasound wave and $z_{1}$ and $z_{2}$ are the acoustic impedances of the materials on either side of the lubricant film. 

The second feature that can also be monitored is the phase change of the reflected signal. Following a similar procedure as before, but this time relating the phase of the reflected wave $\phi_{R}$ to the acoustic properties, yields another equation for the fluid-film thickness,
\begin{equation}
    h = \frac{\rho c^2 (\tan \phi_{R})(z_{2}^2 - z_{1}^2)}{\Omega z_{1} z_{2}^2 \pm \sqrt{(\Omega z_{1} z_{2}^2)^2 - (\tan \phi_{R})^2 (z_{2}^2 - z_{1}^2) (\Omega z_{1} z_{2}^2)^2}}
\label{phase_change_eq}
\end{equation}
Both methods are limited by the range of boundary lengths they can cover. As the thickness increases, the system is said to be no longer stiffness dominated \cite{Beamish2021}, and the reflection coefficient $R$ tends to unity for the amplitude change equation (\ref{amplitude_change_eq}) and $\phi_{R}$ tends to zero for the phase change equation (\ref{phase_change_eq}). The calculated values of $h$ are clearly indefinite at such limits, meaning that these methods are only valid for small boundary lengths. 

The measurable range of fluid-film thickness can be extended by incorporating multiple sensors of varying frequencies, but at the expense of higher hardware complexity in the ultrasonic system.  An alternative to this problem is by means of a third method; the resonant dip method , which essentially involves tracking amplitude \say{dips} that manifest in the frequency domain when the lubricant film resonates under the excitation of the incident wave \cite{pialucha1994}. This method, however, is also limited by its measuring range as higher order dips often blend with background noise.

\begin{figure}
    \centering
    \includegraphics[width=0.5\textwidth]{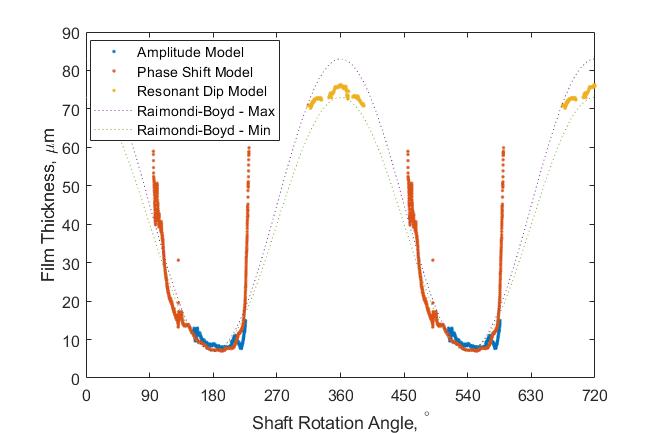}
    \caption{Experimental results where the measurements obtained from the amplitude change model, phase change model and resonant dip technique are displayed.}
    \label{fig:400rpm_20kN_thickness}
\end{figure}

Some of the results obtained from the experimental procedure from \cite{Beamish2021} are repeated here in Figure \ref{fig:400rpm_20kN_thickness}. At first glance, the afore mentioned limitations of the methods are evidenced by gaps in the measurements; the ultrasonic transducer was incapable of capturing enough information in these regions to derive meaningful results. Although the given results qualitatively appear to be in good accordance with the theoretical solution (Raimondi-Boyd method \cite{RaimondiBoyd1958}), these \say{dead zones} might restrict the derivation of other physical effects, such as the film pressure distribution.
Nevertheless, with the given observations, it is still possible to predict what the fluid-film thickness will be at all points around the bearing. One way, would be to assume a deterministic framework and assign the minimum value data point to be the true minimum fluid-film thickness $h_{min}$. The remaining values of $h$ can then be obtained geometrically via,
\begin{equation}
    h = c(1 + \epsilon \cos(\Theta))
\label{fluid_film_thickness_eq}
\end{equation}
where $c$ is the bearing radial clearance, $\Theta$ is a polar coordinate about the shaft-centre, and $\epsilon$ is the eccentricity ratio defined as $\epsilon = \frac{e}{c} = \frac{r-h_{min}}{c}$, with $e$ being the off-set distance of the shaft-centre from the bearing centre.

Assuming a deterministic framework like this implies an absolute certainty that the bearing is perfectly circular, and that the observed experimental results are true, which is unlikely to be the case given the inevitable presence of noise and uncertainties in the measurements. A black-box probabilistic approach was thus adopted here to map a set of observations (shaft rotation angle) to a corresponding vector of targets of the fluid-film thickness measurements. The implementation of a Bayesian approach allows for the prediction of the fluid-film thickness based on the observed data, its noise, uncertainty and any informed beliefs of the underlying physics. In this work, Gaussian Processes (GPs) were chosen as the Bayesian method for regression.

In the following section, a brief background to GPs will be covered. It will then be demonstrated in Section \ref{sec:modelling} how the GPs are constructed to model the fluid-film thickness. Section \ref{sec:results} provides examples of practical applications in which the derived model can be used to determine the shaft-centre position of the bearing. Finally, Section \ref{sec:conclusions} concludes this work with an overview and insights into the proposed method. 

\subsection{Gaussian Process Regression}
\label{sec:GP}
Making predictions in a continuous space of possible input values requires solving a regression problem by fitting a function that maps a dataset of some given observations. More traditional regression techniques involve assuming the underlying characteristics of the fitting function. The caveat of this approach, however, is that such assumptions may lead to poor predictions (e.g.\ linear models attempting to represented a nonlinear trend). 
Gaussian Processes (GPs) \cite{Rasmussen2005} offer powerful predictive capabilities in supervised applications by instead placing a Gaussian prior distribution over an entire function and then weighting all samples that are more likely to fit the observations. Mathematically, this process involves marginalising the function parameters to have the observations conditioned only on the their respective input values and noise. Fundamentally, the GP attempts to model functions of the form,
\begin{equation}
    y = f(x) + \epsilon,   \qquad  \epsilon = \mathcal{N}(0,\sigma_n^2)
\end{equation}
where $y$ is a target value for a given input $x$, $f(x)$ is some function given by the GP and $\epsilon$ is some additive Gaussian noise with zero mean and variance $\sigma_n^2$. The GP is defined by two sets of input points $x$ and $x'$ as,
\begin{equation}
    f(x) \sim \mathcal{GP}(m(x),k(x,x'))
\end{equation}
The form of the function is no longer assumed and, given that it is now defined by a Gaussian, one is left with finding the functions of the mean $m(x)$ and covariance matrix $k(x,x')$. Making predictions for a new test point $y_*$ involves first taking the joint Gaussian distribution over all points of interest,
\begin{equation}
\begin{pmatrix}\mathbf{y}\\
y_*
\end{pmatrix} \sim  \mathcal{N}
\begin{bmatrix}
\begin{pmatrix}
m(X)\\
m(x_*)
\end{pmatrix}\!\!,
\begin{pmatrix}
K(X,X)+\sigma_n^2 & K(X,x_*)\\
K(x_*,X) & K(x_*,x_*)
\end{pmatrix}
\end{bmatrix}\\[2\jot]
\label{eq:joint_gaussian}
\end{equation}
where the new notation $X$ is used to denote a matrix formed of all the $N$ training inputs and $\mathbf{y}$ is the corresponding set of $N$ target observations. Equation (\ref{eq:joint_gaussian}) is a complete representation of the model. However, it is desired to make predictions conditioned on the training set. By recalling the properties of multivariate Gaussian distributions, a subset of variables conditioned on the rest is also distributed as a multivariate Gaussian. Therefore, the conditional distribution over the set of predictions $\mathbf{y^*}$ can be expressed as,
\begin{equation}
    p(\mathbf{y_*}|\mathbf{y}) = \mathcal{N}(\mathbb{E}[\mathbf{y_*}],\mathbb{V}[\mathbf{y_*}])
\end{equation}
where the posterior mean $\mathbf{E}[\mathbf{y_*}]$ and variance $\mathbf{V}[\mathbf{y_*}]$ are defined as,
\begin{equation}
    \mathbb{E}[\mathbf{y_*}] = m(X_*) + K(X_*,X)(K(X,X) + \sigma_n^2\mathcal{I})^{-1}\mathbf{y}
\end{equation}
\begin{equation}
    \mathbb{V}[\mathbf{y_*}] = K(X_*,X_*) - K(X_*,X)(K(X,X) + \sigma_n^2\mathcal{I})^{-1}K(X,X_*)
\end{equation}
As before, the notation $X_*$ and $\mathbf{y_*}$ now includes all the testing points. The covariance (kernel) function must be chosen to fully specify the GP. This function defines the similarity between any two sets of input points, resulting in a symmetric semi-positive square covariance matrix $K$. Defining the covariance matrix in this way ensures that changes in the input space are somewhat correlated to changes in the outputs, providing smooth transitions between neighbouring points. Although a variety of covariance functions exist, choosing the right one could benefit the predictive aptitude of the GP. This consideration will depend on the type of application being undertaken. An example of a commonly-used covariance function would be the \textit{Matérn 3/2} function,
\begin{equation}
    k_{3/2}(x,x') = \sigma_f^2\left(1 + \frac{\sqrt{3}r}{l}\right)\exp{\left(-\frac{\sqrt{3}r}{l}\right)}, \qquad r = ||x - x'||
    \label{eq:matern32}
\end{equation}
\noindent where $\sigma_f^2$ is the signal variance and $l$ represents the length scale. This particular function was chosen as it has been demonstrated to perform well with engineering data for damage localisation \cite{Jones2020}. Since this work will focus on circular geometries, it will be convenient to also introduce covariance functions in the polar domain \cite{Padonou2015}. The similarities of points in Equation (\ref{eq:matern32}) are computed with respect to the Euclidean distance $||x-x'||$, which underestimates the influence of points located on the same concentric circles. A covariance function that satisfies the conditions of semi-positive definiteness over the polar domain is the \textit{$C^2$-Wendland function}, 
\begin{equation}
    W_c(t) = \left( 1 + \tau \frac{t}{c} \right) \left(1 - \frac{t}{c}\right)^{\tau}_{+}, \qquad c \in ] \, 0,\pi ] \,, \tau \geq 4
    \label{eq:wendland}
\end{equation}
where $t$ is the angular distance between points $x$ and $x'$, $\tau$ is a steepening parameter, and $c=\pi$ if the measured distances are geodesic. The conditions in Equation \ref{eq:wendland} are necessary for the covariance to be zero at $t=\pi$ and also strictly positive when $t>\pi$. The influence of the polar kernel can be combined with that of the stationary \textit{Matérn 3/2} function with an ANOVA operation. For two inputs $x$ and $x'$, the final ANOVA combination yields,
\begin{equation}
    k_{2D}(x,x') = s^2(1 + \alpha_1^2 k_{3/2}(\rho,\rho'))(1 + \alpha_2^2 k_{ang}(\theta,\theta'))
    \label{eq:2d}
\end{equation}
where the radial and angular coordinates of $x$ are now represented by $\rho$ and $\theta$, respectively. The polar covariance function is defined by Equation (\ref{eq:wendland}) as $k_{ang}(\theta,\theta') = W_{\pi}(d(\theta,\theta'))$.

The hyperparameters $\sigma_f^2$, $l$, $\tau$, $s^2$, $\alpha_1^2$ and $\alpha_2^2$ in Equations (\ref{eq:matern32}), (\ref{eq:wendland}) and (\ref{eq:2d}) can be determined by maximising the log marginal likelihood of the model. Finding the optimal hyperparamters that maximise the likelihood is equivalent to minimising the negation of the log marginal likelihood. If all the hyperparameters are concatenated into a single vector $\Theta$, then the optimal values are given by,
\begin{equation}
    \hat{\Theta} = \argmin_{\Theta}\{-\log(\mathbf{y}|X,\Theta)\}
\end{equation}
where the log marginal likelihood is defined as \cite{Rasmussen2005},
\begin{equation}
    \log(\mathbf{y}|X,\Theta) = -\frac{1}{2}\mathbf{y}^{T}(K(X,X) + \sigma_n^2\mathcal{I})^{-1}\mathbf{y} - \frac{1}{2}\log\{K(X,X) + \sigma_n^2\mathcal{I}\} - \frac{N}{2}\log(2\pi)
    \label{eq:nlml}
\end{equation}
Equation (\ref{eq:nlml}) can then be evaluated by a numerical optimisation scheme. The \textit{Quantum-Behaved Particle Swarm (QBPS)} optimisation algorithm \cite{Sun2004} was employed to derive the optimised hyperparamters of the covariance functions.

\section{Learning Process}
\label{sec:modelling}

Referring to the mathematical framework presented in Section \ref{sec:GP}, this section will describe the implementation of the proposed modelling process for fluid-film prediction and shaft-centre localisation. The data sets used are identical to those collected in \cite{Beamish2021}, and readers are again referred to their work for a more detailed description of the corresponding experimental procedure. For clarity, however, the experimental setup is briefly reiterated here.
\subsection{Experimental setup}
A bespoke test-rig was constructed to measure the fluid-film thickness of an operating journal bearing (Section \ref{sec:ultrasound_techniques}). Rolling-element bearings were included in the rotor and placed on either side of the journal bearing to adjust the alignment of the shaft. A flexible linkage allowed the rest of the bearing assembly to move freely and adopt its natural position, based on the operating parameters. Controlled variations of the rotational speed, static load and inlet oil temperature were possible. The static load was applied downwards via a hydraulic actuator, and the lubricant was stored in a heated bath as part of a continuous circulating system of a fully-flooded bearing assembly. 
Six longitudinal ultrasonic transducers were embedded within the shaft, feeding to the data acquisition hardware at a rate of $80kHz$ via a multi-channel slip ring. Other measuring equipment were also installed, such as thermocouples to monitor oil temperature, load cells to measure the applied load, encoders to keep record of the rotation angle, and four eddy-current gap sensors for indirect fluid-film thickness measurements.

\subsection{Modelling fluid-film thickness}
For demonstration purposes, measurements obtained from one operational condition only, is first considered. These correspond to the ultrasound measurements gathered while the bearing was operating at a constant speed of $400rpm$ and with an applied static load of $20kN$. The results are those displayed in Figure \ref{fig:400rpm_20kN_thickness}, showing the fluid-film thickness estimates given by the amplitude model, phase model and resonant dip method.

Given the data, it was possible to construct a GP that could best fit the observations. However, it was first necessary to trim the observations in order to remove misleading measurements. The ultrasound techniques used are more reliable in converging regions near the minimum point; cavitation in diverging regions causes the film to rupture, resulting in poor predictions \cite{Kasolang2008}. This effect can be immediately observed in Figure \ref{fig:400rpm_20kN_thickness}, where the phase model predictions converge to unrealistic observations as the sensor readings rotate away from the point of minimum thickness. Similarly, the amplitude model results in an additional dip that challenges the cylindrical geometry of the bearing. Failing to account for these inconsistent observations could negatively affect the predictive capabilities of the GP model. 

\begin{figure}
    \begin{subfigure}[b]{0.49\textwidth}
        \centering
        \captionsetup{justification=centering}
        \includegraphics[width=\textwidth]{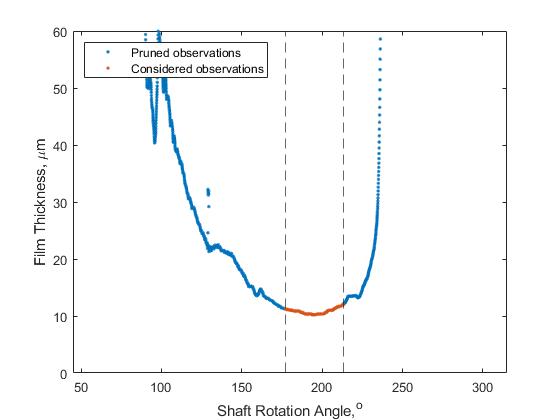} 
        \caption{}
        \label{fig:data set_phase}
    \end{subfigure}
    \hfill
    \begin{subfigure}[b]{0.49\textwidth}
        \centering
        \captionsetup{justification=centering}
        \includegraphics[width=\textwidth]{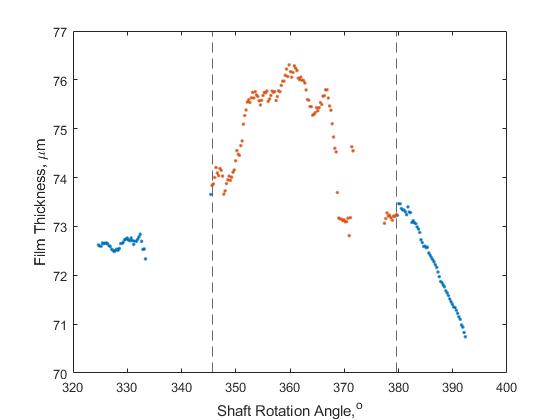} 
        \caption{}
        \label{fig:dataset_resdip}
    \end{subfigure}
\caption{Data selection process for modelling.}
\label{fig:dataset_construction}
\end{figure}

For this study, only data contained within a small region around the minimum fluid-film thickness was considered. The region was chosen to span $10\%$ of a full cycle; that is, $18^o$ on either side of the minimum point (Figure \ref{fig:dataset_construction}). The Raimondi-Boyd (min) solution was used as a reference to estimate the angular position of the minimum fluid-film thickness. Similarly, the procedure was followed for the observations given by the resonant dip technique. A data set was then constructed by combining the observations from the phase change method with those from the resonant dip method before partitioning it into a training and testing set. In this case, the data selection process is based on a predefined range, possibly delimiting the most consistent observations. This parameter will influence the learning process and a form of validation is required to ensure that the considered portion maximises the performance of the model. Such a validation procedure is not developed in this paper; it will be pursued in future work. Nevertheless, it will be demonstrated that meaningful results are still achieved, that highlight the intended ideas of this section.

Although the Matérn 3/2 function (equation (\ref{eq:matern32})) generalises well in engineering applications, it is worth evaluating other covariance functions that might better serve the current application. In this case, three covariance functions were used to construct each corresponding GP: (a) a squared-exponential function, (b) a Matérn 3/2 function and (c) a strictly-periodic function. The squared-exponential and periodic functions are given by, 
\begin{equation}
    k_{sqe}(x,x') = \sigma_f^2\left( \frac{r^2}{2l^2}\right), \qquad r = ||x - x'||
    \label{eq:sqe}
\end{equation}
and,
\begin{equation}
    k_p(x,x') = \sigma_f^2\exp\left(-2\frac{(\pi|x -x'|/p)}{l^2} \right)
    \label{eq:periodic}
\end{equation}
respectively. As in the Matérn function, the parameters $\sigma_f^2$ and $l$ represent the signal variance and length scale, respectively, and the new parameter $p$ in equation (\ref{eq:periodic}) simply determines the period of the function.

\begin{figure}
    \begin{subfigure}[b]{0.32\textwidth}
        \centering
        \captionsetup{justification=centering}
        \includegraphics[width=\textwidth]{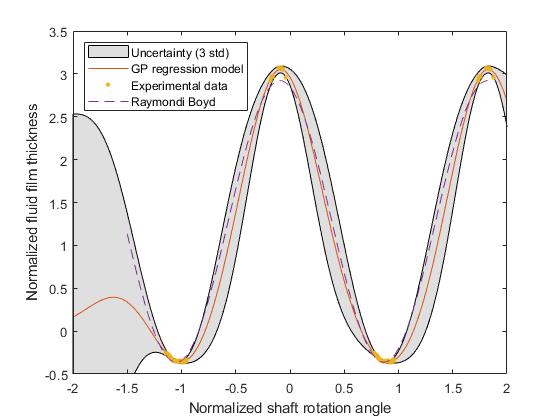} 
        \caption{}
    \end{subfigure}
    \hfill
    \begin{subfigure}[b]{0.32\textwidth}
        \centering
        \captionsetup{justification=centering}
        \includegraphics[width=\textwidth]{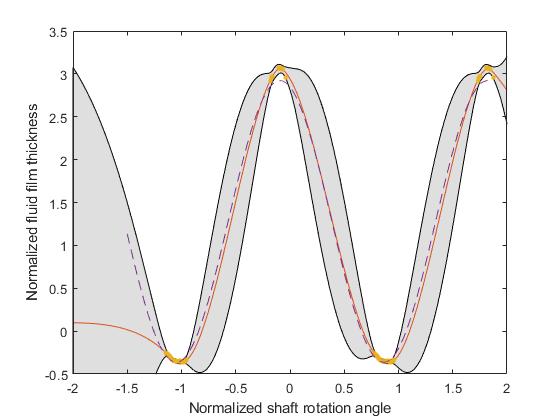} 
        \caption{}
    \end{subfigure}
    \hfill
    \begin{subfigure}[b]{0.32\textwidth}
        \centering
        \captionsetup{justification=centering}
        \includegraphics[width=\textwidth]{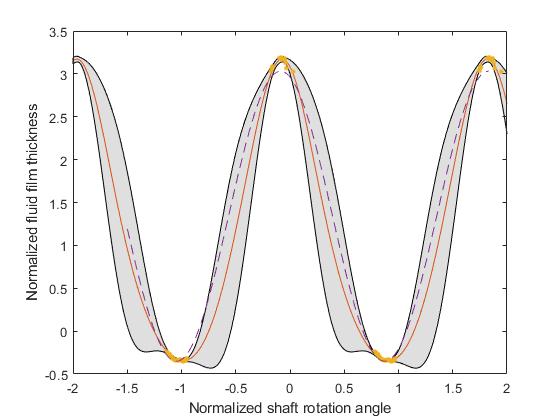}
        \caption{}
    \end{subfigure}
\caption{GPs over phase change fluid-film thickness observations defined by the (a) squared exponential function, (b) Matérn $3/2$ function and (c) strictly periodic function. Log marginal likelihood evaluated on test data resulted in (a) 583.38, (b) 582.43 and (c) 597.97 for each model, respectively.}
\label{fig:gp_film_thickness}
\end{figure}

The resultant models for each data set can be observed in Figure \ref{fig:gp_film_thickness}. Unlike the former two functions, the periodic function retains the periodicity of the GP away from the observations. This effect is in accordance with the physical properties of the bearing operating at a constant speed. Additionally, after evaluating the log marginal likelihood of the testing sets with respect to each GP, the periodic prior appears to be the one that best supports the new evidence. Given these advantages, the GPs constructed for all other operational conditions were defined by the periodic covariance function.
It is worth noting that the GP interpolates the film measurements over the full range of the bearing. All three GPs appear to contain the Raimondi-Boyd solution within their delimited uncertainties. This interpolation is not based on the observations of the minimum fluid-film thickness only, but it also takes into consideration all the other observations in the training set. By making use of the available measurements, the GP can generate a probabilistic prediction of the actual film thickness for all shaft rotation angles. It is important to notice the uncertainty inferred by the model; the prediction becomes more uncertain in areas where observations were unavailable. This result comes as no surprise, given that the model outputs a prediction in these locations without empirical evidence. Nevertheless, the GP posterior mean indicates where observations are most likely to exist, and one could thus use the Maximum-a-Posteriori (MAP) estimate to model the fluid-film thickness. Alternatively, the uncertainty in the predictions can be propagated through further analysis to estimate some of the bearing dynamics with a quantified measure of confidence.

In a condition monitoring context, for example, having a GP learn from measurements corresponding to a healthy bearing, could indicate whether new observations actually derive from a normal operating condition. If wear or misalignment were to emerge during operation, the resultant observations would somewhat deviate from normal, and their likelihood with respect to the already inferred model would reduce as a result. 

For now, the focus of this work is on predictions near the minimum fluid-film thickness, and the potential applications of the presented strategies for condition monitoring will be left for future work. The following section demonstrates a similar probabilistic reasoning to estimate the shaft-centre location.

\subsection{Pre-processing for shaft-centre localisation strategy}
The shaft-centre location can be defined by two coordinates: its eccentricity, $e$, and attitude angle, $\Phi$. These coordinates represent the radial and angular components of the shaft-centre location with respect to the absolute bearing centre and load vector, respectively. It is possible to extract these coordinates directly from the fluid-film thickness observations. This approach involves finding the minimum value of the observations and its corresponding polar coordinate with respect to the bearing's \textit{Top Dead Centre} (TDC). However, one would assume a deterministic framework by deeming the minimum point observation to be true. As mentioned earlier in Section \ref{sec:ultrasound_techniques}, assuming absolute certainty in the observations would imply perfection and thus lead to inaccurate predictions of the shaft-centre location. Therefore, the probabilistic framework described in the previous section was preferred to establish the most likely location of the shaft when given a specific operational condition.

\begin{figure}
    \centering
    \includegraphics[width=0.32\textwidth]{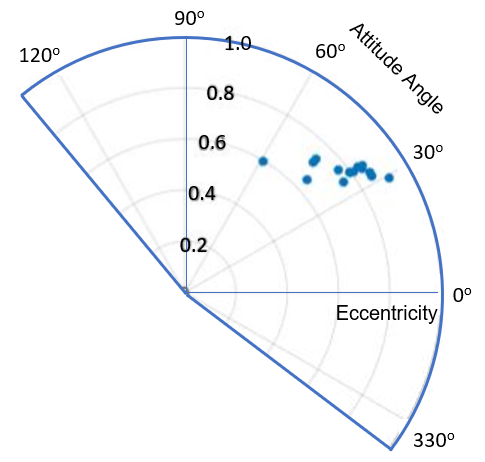}
    \caption{Training data set including all shaft-centre locations.}
    \label{fig:data_set}
\end{figure}

The experimental runs involved either changing the applied load $W$ or rotational speed $\omega$ in controlled steady-state environments (e.g.\ constant oil temperate and supply). The modelling process described in Section \ref{sec:modelling} was applied to each of these data sets. From the generated GPs, it is then possible to extract the MAP estimation of $h_{min}$ and its corresponding angular position in terms of the attitude angle. Figure \ref{fig:data_set} includes the coordinates of the shaft-centre inferred from each speed-load combination. Although unusual, the shaft found itself at the top-half of the bearing throughout the experimental procedure. This was expected, given that the bearing assembly adjusts to the build-up of pressure produced in the film by reacting to the load being applied downwards on the housing (while maintaining the shaft fixed). This setup is equivalent to having the bearing assembly fixed and the loaded shaft free to adjust to its stable position.
At this point, it was necessary to assign labels to each location in order to construct a data set that could be used to map the shaft-centre location from its respective operational parameters. A convenient combination of these parameters exists via the Sommerfeld number \cite{Hori2006}, defined by,
\begin{equation}
    S_o = \left( \frac{R}{c} \right)^2\frac{\mu\omega LR}{2W}
    \label{eq:sommerfeld}
\end{equation}
where $R$ and $L$ are the internal radius and length of the bearing, respectively, and $\mu$ is the lubricant dynamic viscosity. The Sommerfeld number has been commonly used to compare the performance of journal bearings \cite{He2016}. It should be noted that the rotational speed and applied load variables in equation (\ref{eq:sommerfeld}) are combined by the ratio $\omega/W$. Given that the experimental procedure was performed in a controlled steady-state environment, the viscosity term is assumed constant. Hence, each location can be labelled by a simplified Sommerfeld number defined only as a proportion of their corresponding speed-load ratios. This assumption allows one to have a training set $D = \left\{ X_i,y_i \right\}^N_{i=1}$ of $N$ observations where $X_i$ is the coordinate $( \rho_i,\theta_i )$ of the target $y_i = \omega_i / W_i$. Since the bearing is analysed in the polar domain, the coordinates are expressed in terms of a radial distance $\rho_i$ and angular position $\theta_i$ from the bearing's centre and TDC, respectively.

\subsection{Shaft-centre localisation with Gaussian Processes}
\label{sec:results}

Having established a data set that relates the shaft-centre position to the corresponding operating parameters, it is possible to develop a new GP regression model aiming to capture this relationship at all locations the shaft might adopt. This procedure involves placing a grid over the bearing's cross-section. The nodes in the grid correspond to the predicted values of the model based on the training set $D$ and any informed beliefs of the function being modelled. These considerations are managed with a suitable choice of covariance (kernel) functions that best describe the underlying behaviour of the journal bearing. 

The polar nature of the bearing analysis means that a unique Euclidean-based covariance function would underestimate the influence of changes in the angular direction and disregard the circular geometry of the bearing. A more prudent approach was instead followed with the inclusion of a polar covariance function $k_{ang}$ defined by the $C^2$-Wendland function (Equation (\ref{eq:wendland})) for geodesic distances. Two other covariance functions were also considered for the radial space and then combined with $k_{ang}$ via an ANOVA operation; these were the polynomial exponential decay function ($k_{pol}$) and the Mat\'{e}rn $3/2$ function ($k_{3/2}$), and each combination will be referred to as \textit{Model A} and \textit{Model B}, respectively. 

\begin{figure}
    \begin{subfigure}[b]{0.32\textwidth}
        \centering
        \captionsetup{justification=centering}
        \includegraphics[width=\textwidth]{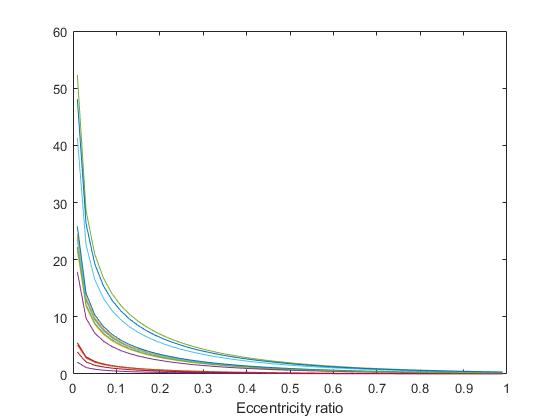} 
        \caption{}
    \end{subfigure}
    \hfill
    \begin{subfigure}[b]{0.32\textwidth}
        \centering
        \captionsetup{justification=centering}
        \includegraphics[width=\textwidth]{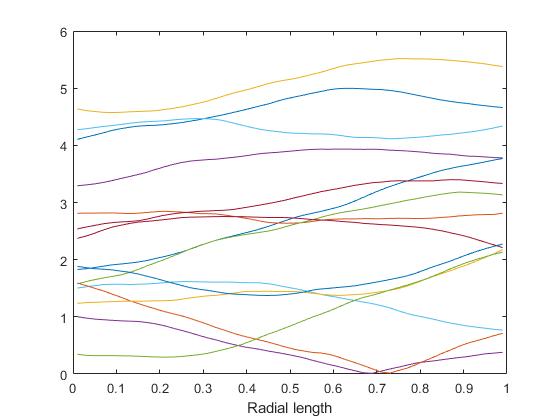} 
        \caption{}
    \end{subfigure}
    \hfill
    \begin{subfigure}[b]{0.32\textwidth}
        \centering
        \captionsetup{justification=centering}
        \includegraphics[width=\textwidth]{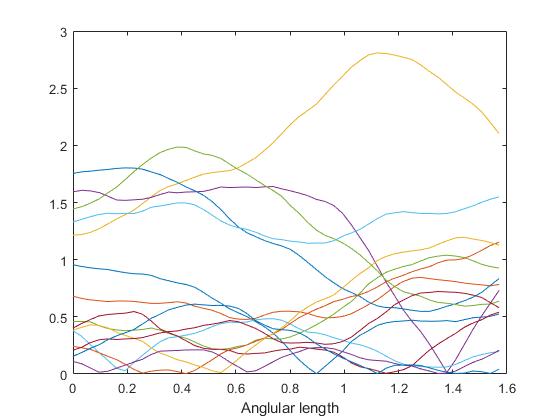}
        \caption{}
    \end{subfigure}
\caption{Samples drawn from zero-mean GP priors defined by: (a) a polynomial exponential-decay kernel, (b) a Matérn $3/2$ kernel, and (c) $C^2$-Wendland kernel. Each line represents a different random function sampled from the GP prior. The vertical axis represents the output of these functions when presented with values from the input space.}
\label{fig:prior_samples}
\end{figure}

\begin{figure}
    \begin{subfigure}[b]{0.32\textwidth}
        \centering
        \captionsetup{justification=centering}
        \includegraphics[width=\textwidth]{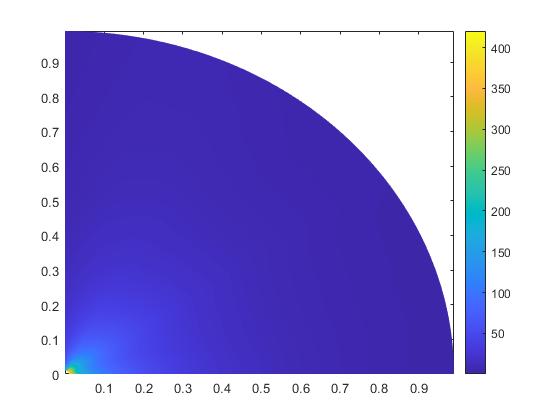} 
        \caption{}
        \label{fig:poly_prior}
    \end{subfigure}
    \hfill
    \begin{subfigure}[b]{0.32\textwidth}
        \centering
        \captionsetup{justification=centering}
        \includegraphics[width=\textwidth]{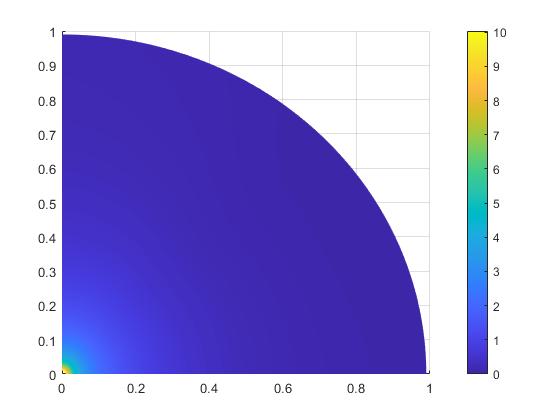} 
        \caption{}
        \label{fig:poly_post}
    \end{subfigure}
    \hfill
    \begin{subfigure}[b]{0.32\textwidth}
        \centering
        \captionsetup{justification=centering}
        \includegraphics[width=\textwidth]{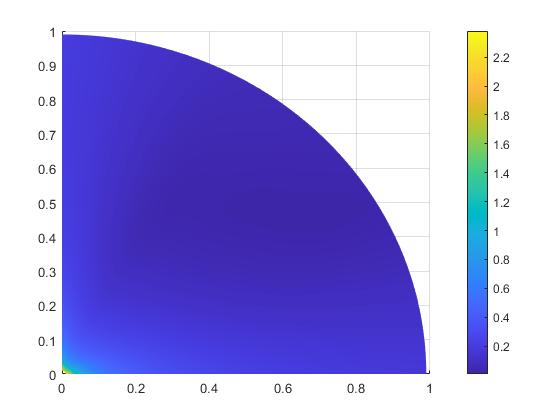}
        \caption{}
        \label{fig:poly_var}
    \end{subfigure}
    \begin{subfigure}[b]{0.32\textwidth}
        \centering
        \captionsetup{justification=centering}
        \includegraphics[width=\textwidth]{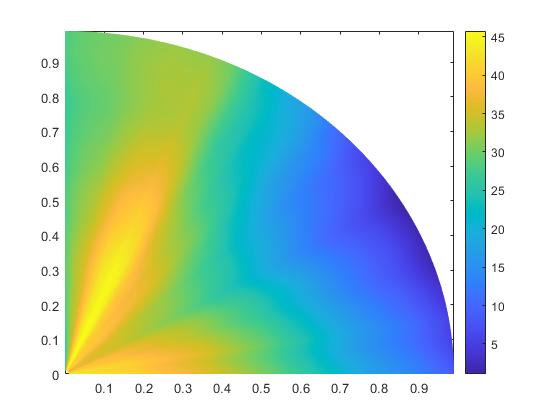} 
        \caption{}
        \label{fig:matern_prior}
    \end{subfigure}
    \hfill
    \begin{subfigure}[b]{0.32\textwidth}
        \centering
        \captionsetup{justification=centering}
        \includegraphics[width=\textwidth]{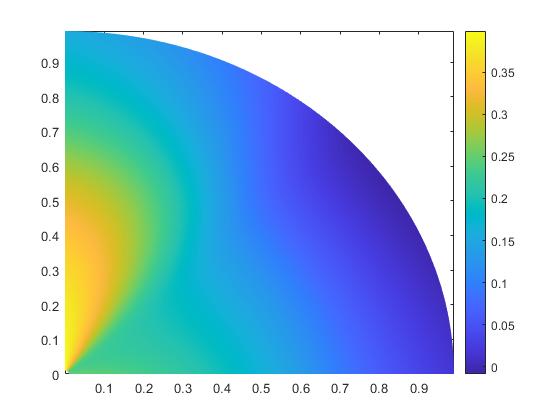} 
        \caption{}
        \label{fig:matern_post}
    \end{subfigure}
    \hfill
    \begin{subfigure}[b]{0.32\textwidth}
        \centering
        \captionsetup{justification=centering}
        \includegraphics[width=\textwidth]{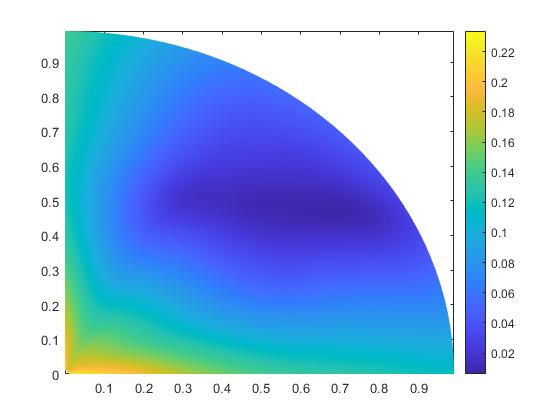}
        \caption{}
        \label{fig:matern_var}
    \end{subfigure}
\caption{(a)-(c) GP prior sample, GP posterior mean and variance defined by Model A. (d)-(f) GP prior sample, GP posterior mean and variance defined by Model B.}
\label{fig:2D_models}
\end{figure}

\begin{figure}
    \begin{subfigure}[b]{0.48\textwidth}
        \centering
        \captionsetup{justification=centering}
        \includegraphics[width=\textwidth]{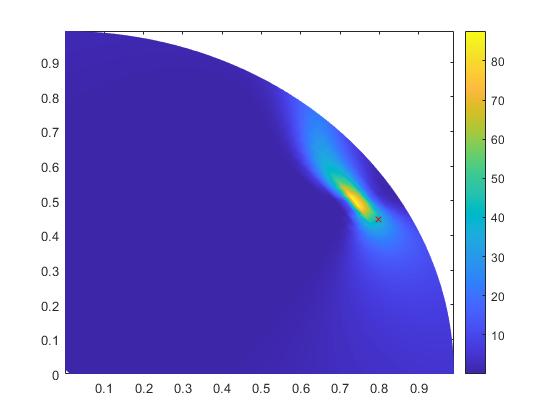} 
        \caption{}
    \end{subfigure}
    \hfill
    \begin{subfigure}[b]{0.48\textwidth}
        \centering
        \captionsetup{justification=centering}
        \includegraphics[width=\textwidth]{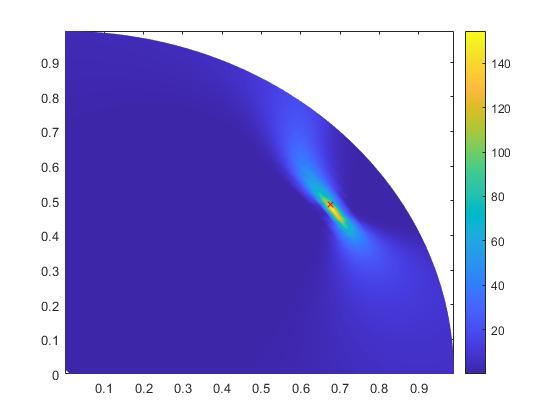} 
        \caption{}
    \end{subfigure}
    \hfill
    \begin{subfigure}[b]{0.48\textwidth}
        \centering
        \captionsetup{justification=centering}
        \includegraphics[width=\textwidth]{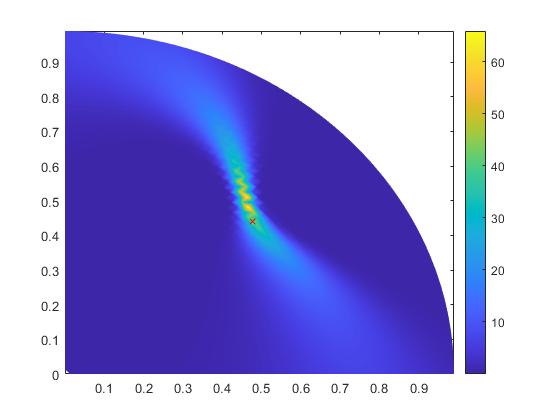}
        \caption{}
    \end{subfigure}
    \hfill
    \begin{subfigure}[b]{0.48\textwidth}
        \centering
        \captionsetup{justification=centering}
        \includegraphics[width=\textwidth]{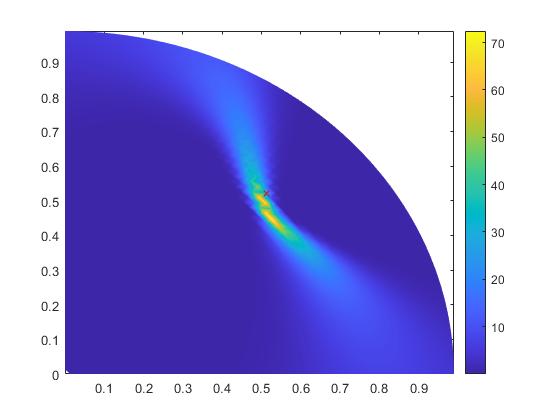}
        \caption{}
    \end{subfigure}    
\caption{\textbf{Model A} likelihood maps of bearing shaft-centre location. Each figure corresponds to an arbitrary test-point: (a) $100rpm/10kN$, (b) $400rpm/14kN$, (c) $400rpm/4kN$ and  (d) $800rpm/10kN$. The actual measurements are indicated by a red cross.}
\label{fig:likeli_maps_poly}
\end{figure}

\begin{figure}
    \begin{subfigure}[b]{0.48\textwidth}
        \centering
        \captionsetup{justification=centering}
        \includegraphics[width=\textwidth]{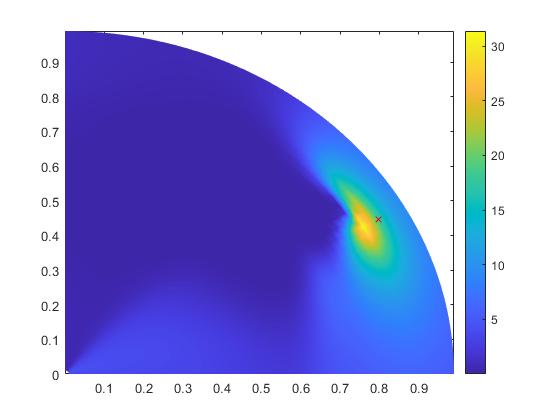} 
        \caption{}
    \end{subfigure}
    \hfill
    \begin{subfigure}[b]{0.48\textwidth}
        \centering
        \captionsetup{justification=centering}
        \includegraphics[width=\textwidth]{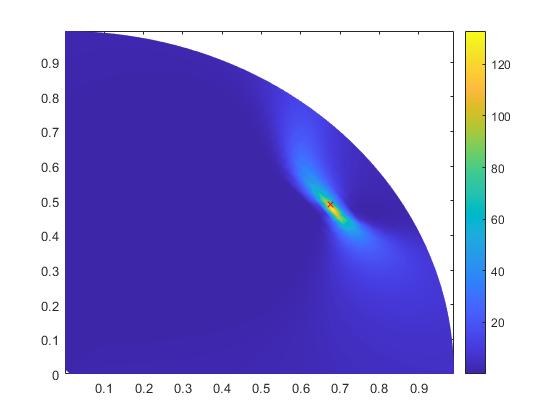} 
        \caption{}
    \end{subfigure}
    \hfill
    \begin{subfigure}[b]{0.48\textwidth}
        \centering
        \captionsetup{justification=centering}
        \includegraphics[width=\textwidth]{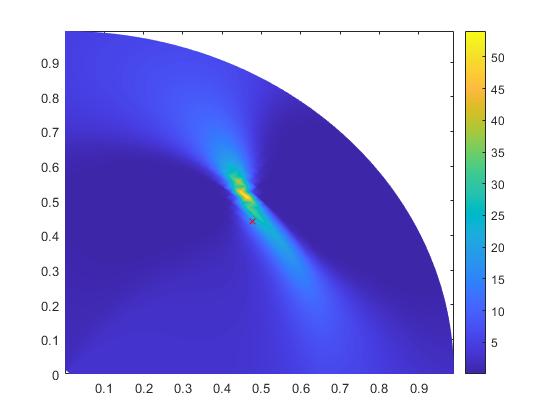}
        \caption{}
    \end{subfigure}
    \hfill
    \begin{subfigure}[b]{0.48\textwidth}
        \centering
        \captionsetup{justification=centering}
        \includegraphics[width=\textwidth]{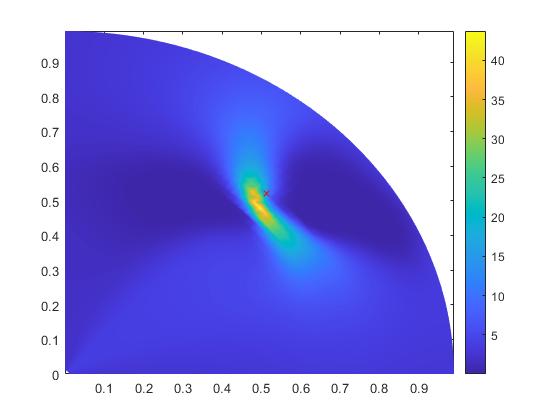}
        \caption{}
    \end{subfigure}    
\caption{\textbf{Model B} likelihood maps of bearing shaft-centre location. Each figure corresponds to an arbitrary test-point: (a) $100rpm/10kN$, (b) $400rpm/14kN$, (c) $400rpm/4kN$ and (d) $800rpm/10kN$. The actual measurements are indicated by a red cross.}
\label{fig:likeli_maps_matern}
\end{figure}

The former radial covariance function ($k_{pol}$) is justified when considering the expected behaviour of the journal bearing under varying operational conditions. Increasing the rotational speed results in the journal adjusting towards the centre of the bearing. In theory, given a steady-state fixed load operation, the journal will be concentric to the bearing when $y \propto \omega \to \infty$ as $\omega \to \infty$. Conversely, increasing the load while maintaining a constant speed will have an opposite effect and $y \propto 1/W \to 0$ as $W \to \infty$. An exponential decay function will somewhat model these characteristics and is thereby expected to provide a more realistic predictive framework. Figure \ref{fig:prior_samples} shows some random samples from GP priors defined by each of the individual covariance functions. Having defined the priors, the mean and variance of the GP posterior can now be inferred by introducing the training set. The results are presented in Figure \ref{fig:2D_models}, showing both the mean and variance of GP prior and posterior corresponding to each model. Only the first quadrant of the bearing's cross-section was modelled since the shaft was found to have positioned itself only within this area in all experiments. This choice was also convenient to minimise computational expenses. Much like in the previous section, the models (more evident in Model B) are less confident in regions away from the observations.
The predictive distribution is not directly estimating the location of the journal centre. It would be more convenient to instead predict the journal position given a new combination of operational parameters. Fortunately, having assigned speed-load ratio predictions to each grid point $(\rho,\theta)$, it is possible to determine a coordinate $X$ that maximises the probability of observing a new speed-load ratio $y_{new}$. Inverting the problem in this way means assessing the log-likelihood of a new observation, $y_{new}$, as,
\begin{equation}
    \log p(y_{new}|D,X_{*},y_{*}) = -\frac{1}{2}\log\mathbb{V}[y_{*}] - \frac{(y_{new} - \mathbb{E}[y_{*}])^2}{2\mathbb{V}[y_{*}]} - \frac{1}{2}\log2\pi
    \label{eq:log_likelihood}
\end{equation}
where $X_{*}$ is an array comprised of all the candidate coordinates where the log-likelihood of $y_{new}$ is evaluated. The result can be visualised in Figures \ref{fig:likeli_maps_poly} and \ref{fig:likeli_maps_matern}, displaying the likelihood map of $y_{new}$ at all grid points across the bearing section. One can thus infer the journal position by locating the point of maximum likelihood in the grid, providing also a measure of uncertainty in the prediction.

It is important to note that the model simply attempts to predict the mean position of the shaft-centre, given the initially-presented observations. In this case study, there is no reason to believe that any form of elliptical orbiting occurred. However, if some form of elliptical motion exists, and enough observations are provided, then the GP model should account for the added uncertainty given by the variation of the shaft displacement about its centre. If this were to be the case, then the estimated location should indicate the area covered by the orbital motion of the shaft-centre, but predicting the orbital path may require a different approach in the learning process.

Another point that should be mentioned, is the limitation of having the rotational speed and static load as the only two input parameters of the model. The journal bearing was studied in a controlled environment, and it would be unrealistic to expect an identical behaviour in a practical application. In order to improve the robustness of the GP model, one would need to consider other important parameters, such as dynamic loads, lubricant temperature variations, elastic deformations and surface roughness, among others. This work leaves room for these considerations and will be accounted for in future experiments.


\section{Performance Evaluation}

Because of the small available data set, splitting the observations into a training and test set could have limited the learning process. Nevertheless, alternative validations techniques exist to account for small data sets, such as \textit{k-fold cross-validation} and \textit{leave-one-out cross-validation} (LOOCV) \cite{Molinaro2005}. Given that the number of observations was particularly small in this work ($N=15$), the latter approach was adopted. 

The LOOCV approach involves splitting the whole data set into $K$ parts equal to the total number of observations (i.e. $K=N$).  The model is then trained on $K-1$ parts and tested with the remaining one. This process is repeated with a newly assigned test point until all $K$ parts have been a test set. The calculated errors in each iteration are then averaged to give a final performance evaluation. Although training the model repetitively can be computationally expensive, a small data set compensates for this inconvenience. 

The localisation algorithms developed in the previous section were based on different prior beliefs. The small number of observation makes the choice of a suitable GP prior even more important. One can make use of the marginal likelihood to compare how likely unobserved data ($y_{new}$) are given the choice of prior belief. The errors in the validation process were hence assessed by calculating the likelihood of each test set with Equation (\ref{eq:log_likelihood}). The results (Table \ref{tab:perf_eval}) demonstrate that the more informed prior leads to a higher overall likelihood of the test data. 

\begin{figure}
    \begin{subfigure}[b]{0.48\textwidth}
        \centering
        \captionsetup{justification=centering}
        \includegraphics[width=\textwidth]{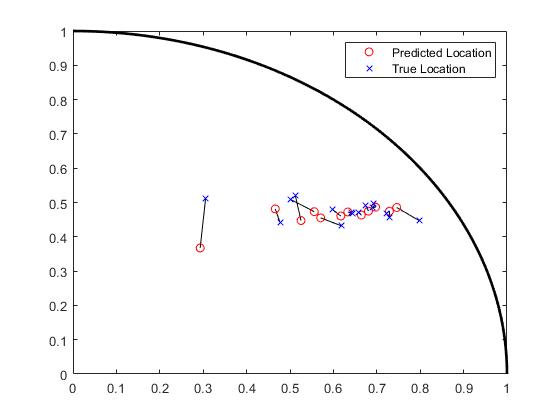} 
        \caption{$k_{pol}(\rho,\rho')$ \& $k_{ang}(\theta,\theta')$}
    \end{subfigure}
    \hfill
    \begin{subfigure}[b]{0.48\textwidth}
        \centering
        \captionsetup{justification=centering}
        \includegraphics[width=\textwidth]{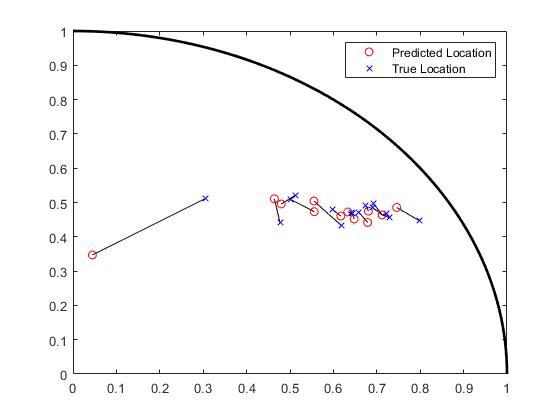} 
        \caption{$k_{3/2}(\rho,\rho')$ \& $k_{ang}(\theta,\theta')$}
    \end{subfigure}
\caption{Distance of the predicted locations to their corresponding true values. (a) Model A and (b) Model B.}
\label{fig:rms_results}
\end{figure}

Another performance assessment method was also considered where the most likely predicted location was compared to the true value via the \textit{Root-Mean-Square-Error} (RMSE). As before, the RMSE was evaluated at each iteration test set of the LOOCV process and then averaged to attain the overall performance. The results have also been included in Table \ref{tab:perf_eval} and a depiction of these differences can be seen in Figure \ref{fig:rms_results}. 
\begin{table}
\centering
\caption{Performance evaluation results of GP models indicating the averaged marginal likelihood and root-mean-square-error (RMSE). The results are averages of the error metrics derived from the test sets assigned at each iteration in the cross-validation process. }
\label{tab:perf_eval}
\resizebox{\textwidth}{!}{%
\begin{tabular}{cccc}
\hline
Model & Covariance function                                & Averaged marginal likelihood & Averaged RMSE \\ \hline
A     & $k_{pol}(\rho,\rho')$ \& $k_{ang}(\theta,\theta')$ & 70.8083                      & 0.0442        \\
B     & $k_{3/2}(\rho,\rho')$ \& $k_{ang}(\theta,\theta')$ & 50.4647                      & 0.0685        \\ \hline
\end{tabular}%
}
\end{table}
Similarly, the averaged RMSE indicates that Model A predicted locations that were generally closer to the true values than Model B. By looking at the predicted locations in Figure \ref{fig:rms_results}, one can note that a major contribution to this error is given by the point lying nearest to the bearing centre. This point corresponds to the journal bearing operating at a speed of $400rpm$ and with an applied static load of $2kN$. Under these conditions, the shaft is located away from all the other locations recorded in the dataset. In areas of sparse data, the GP will be less certain and give poorer predictions, like that observed for this data point. However, when considering the probabilistic distribution over the shaft-centre location, the true location is said to exist somewhere within the spread of the prediction. That is, the GP predictions are not limited to a single location and account for the probabilities of the true values existing anywhere in the bearing. The practical advantage of this probabilistic framework is then provided by reduced areas within the bearing where the shaft-centre true location is more likely to be found. These areas only become narrower (i.e. more certain) under the presence of data, and more measurements would be needed to improve the performance of the proposed models.

\section{Conclusions}
\label{sec:conclusions}

The presented work can be divided into two parts. The first described the construction of a GP model over fluid-film thickness measurements. It covered how ultrasound methods can be limited by the range of film thickness they can measure, leaving \say{dead zones} in the observations. Treating the problem as a form of regression allowed for the prediction of film thickness values over these \say{dead zones}, and thereby covered the full range of the journal bearing. The attained results could have the potential to derive other hydrodynamic parameters, such as the pressure profile along the film. Additionally, the inferred model could allow for condition monitoring strategies in the detection unwanted operations caused by surface wear or misalignment. In this case, however, the GPs were used to locate the bearing shaft under various operational conditions, which were then aggregated into a new data set. 

The second part proposed using the new data set to construct a likelihood map capable of predicting the location of a journal bearing shaft-centre for any given operational condition. That is, by modelling a discrete grid over the bearing's bore, the location of the shaft-centre can be quantified at all possible points across the grid, and the point returning the maximum likelihood will correspond to the most likely location of the shaft within the bearing. This approach offers the possibility to intuitively observe the confidence of the predictions. In other words, it does not return a single point with absolute certainty and instead allows the true location to be found within an area of high probability in the bearing's bore. 

This form of visualisation could assist the design process of a journal bearing. If enough data points are provided, then a model like this could return the location of the shaft-centre without the need for complex oil film measurement systems. Having a fully-trained model, it could also be transferred and applied to other bearings of similar characteristics. 

Nevertheless, the possibilities of the work presented in this paper still needs to be further investigated to ensure the model is capable of reliably returning accurate predictions. Future work also includes evaluating other operational parameters in the prediction of the shaft location, such as temperature changes and response to dynamic loads.

\vspace{24pt}

\noindent \textbf{Acknowledgements} \vspace{12pt}

The authors gratefully acknowledge the support of the UK Engineering and Physical Sciences Research Council (EPSRC) through Grant reference EP/R004900/1.

\renewcommand{\bibname}{References}

\bibliographystyle{unsrt}
\bibliography{refs}

\end{document}